\documentclass[letterpaper, 10 pt, conference]{ieeeconf}  
\usepackage{soul} 

\IEEEoverridecommandlockouts                              

\usepackage{booktabs}

\usepackage{graphicx}
\usepackage{float}    
\begin{document}

\title{\LARGE \bf
ExFace: Expressive Facial Control for Humanoid Robots \\with Diffusion Transformers and Bootstrap Training
}

\author{Dong Zhang$^{*}$,~Jingwei Peng$^{*}$,~Yuyang Jiao,~Jiayuan Gu,~Jingyi Yu,~Jiahao Chen$^\dagger$
\thanks{This work was supported in part by Shanghai Frontiers Science Center of Human-centered Artificial Intelligence, and the MoE Key Lab of Intelligent Perception and Human-Machine Collaboration (ShanghaiTech University).
}
\thanks{The authors are with the School of Information Science and Technology, ShanghaiTech University, Shanghai 201210, China.}
\thanks{$~^\ast$ Equal contributors.}%
\thanks{$~^\dagger$ Corresponding author
{\tt\small chenjh2@shanghaitech.edu.cn}}%
}
\maketitle
\thispagestyle{empty}
\pagestyle{empty}

\begin{abstract}

This paper presents a novel Expressive Facial Control (ExFace) method based on Diffusion Transformers, which achieves precise mapping from human facial blendshapes to bionic robot motor control. By incorporating an innovative model bootstrap training strategy, our approach not only generates high-quality facial expressions but also significantly improves accuracy and smoothness. Experimental results demonstrate that the proposed method outperforms previous methods in terms of accuracy, frame per second (FPS), and response time. Furthermore, we develop the ExFace dataset driven by human facial data. ExFace shows excellent real-time performance and natural expression rendering in applications such as robot performances and human-robot interactions, offering a new solution for bionic robot interaction.

\end{abstract}


\section{Introduction}
Facial expressions are integral to human communication, playing a pivotal role in the transmission of emotions, attitudes, and intentions. As evidenced in prior research, individuals rely on a variety of facial expressions to both convey and interpret affective states~\cite{plutchik2014emotions}. The ability to recognize and generate facial expressions allows individuals to navigate complex social interactions by understanding the emotions and intentions of others~\cite{fong2003survey}. Therefore, the development of robotic systems capable of autonomously replicating a wide range of human facial expressions is essential for advancing human-robot interaction. This capability is particularly significant in fostering more natural and effective communication between humans and machines~\cite{reissland1988neonatal}. Moreover, the ability of robots to mimic human facial expressions is crucial for enhancing their emotional responsiveness and interactional quality. This process is key to improving the emotional intelligence of robots, making them more effective in roles such as caregiving, education, and entertainment~\cite{hakli2014social}.



Traditional methods~\cite{ lin2016expressional} for robot mimicry of human facial expressions require human expertise and cannot account for the full variability of human facial expressions~\cite{fong2003survey,blow2006art}. 
Learning-based approaches~\cite{chen2021smile,wu2024retargeting} leverage neural networks to learn complex mappings between facial expressions, e.g., represented in human blendshape~\cite{joshi2006learning} data, and motor control signals, enabling robots to mimic human expressions with greater accuracy and adaptability.
However, these methods face notable challenges. For example, convolutional networks can struggle to effectively capture multi-modal distributions~\cite{chen2021smile}, while others suffer from slow processing speeds and low frame rates~\cite{wu2024retargeting}. 
Additionally, these methods are often restricted to processing single-frame data, preventing them from handling the dynamic, continuous nature of human facial expressions, which is crucial for accurate and expressive emotional communication. This limitation often results in noticeable jitter and unnatural movements in practical applications.

\begin{figure}[!t]
\centering
\includegraphics[width=3.3in]{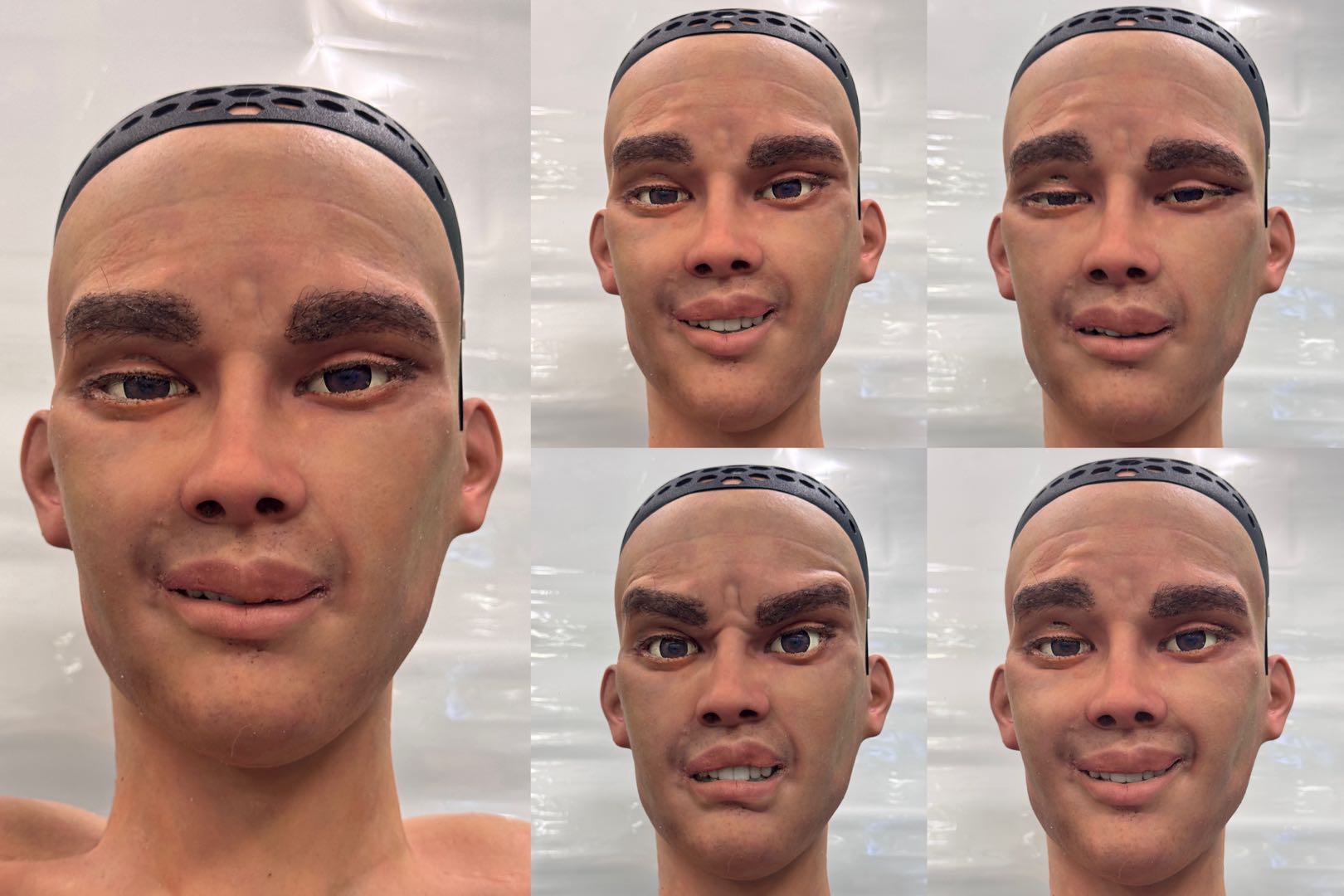}
\caption{Micheal, the realistic humanoid robot, demonstrates its interactive capabilities by displaying a variety of facial expressions, with these motions driven by ExFace. Its dynamic range of expressions highlights the potential for natural, human-like interaction between robots and people.}
\label{fig_1}
\end{figure}

To address these challenges, we propose \textbf{ExFace}, a novel system featuring Diffusion Transformers~\cite{peebles2023scalable}. By leveraging the strengths of both diffusion models~\cite{ho2020denoising} and Transformer~\cite{vaswani2017attention} architectures, ExFace efficiently maps dynamic human blendshape sequences to robot motor signals in a generative way, enabling precise, smooth, and expressive robotic facial control.
Besides, the Transformer architecture provides a unified approach that supports both single-frame and sequential predictions. Specifically, we integrate historical frames with the current frame during real-time inference, to generate the current desired motor command extracted from the output sequence. It distinguishes our method from previous works by fully leveraging sequential information, enabling the system to handle complex, dynamic facial expressions in real time and generate more coherent outputs.

Furthermore, we introduce an innovative bootstrap training strategy in which the model iteratively improves itself using the data it generates. Specifically, we start by applying random single-frame motor signals to collect static robot blendshape data, establishing an initial mapping between blendshapes and motor signals. Next, we use this preliminary model to control the robot, enabling it to collect dynamic robot blendshape sequences in response to human blendshape input. By continuously gathering new data and fine-tuning on it, the model progressively refines its performance, enhancing the accuracy of expression mapping and enabling more seamless and natural robotic facial control.

Experimental results demonstrate that ExFace outperforms prior methods, by delivering real-time, smooth, and natural robotic facial expressions. The output control commands are rendered at 60 fps, ensuring smooth and realistic facial movements. We have improved the system's response time, with control commands being executed within 0.15 seconds from detecting a human expression to the robot's action.

In addition, our approach demonstrates remarkable adaptability. We successfully validated this method on two different robots: Micheal shown in Fig.\ref{fig_1}, a cable-driven robot that uses cable pulls for movement, and Hobbs, a robot that uses articulated linkages for control. To further support future research, we have released the ExFace dataset, which includes robot facial images, motor control values, and corresponding blendshape data.



\section{Related works}

Early facial retargeting approaches relied on pre-programmed expression templates, defining a set of basic expressions in 3D space and using interpolation techniques to generate various expressions. For example, the Kismet system~\cite{scherer1992does} interpolated among basic facial postures to create diverse expressions, while Albert HuBo~\cite{oh2006design} mimicked specific human actions to replicate facial expressions. Similarly, Asheber~\cite{asheber2016humanoid} proposed a simplified design to enhance flexibility and adaptability in expression transformation. However, such template-based methods are limited by fixed expression sets and require extensive manual intervention, resulting in low efficiency and scalability.

The introduction of neural networks marked a significant shift in the field, as deep learning techniques began to be applied to facial expression driving. For instance, Chen et al.~\cite{chen2021smile} utilized facial landmark recognition to learn the mapping between landmarks and servo control values, while Wu et al.~\cite{wu2024retargeting} employed optimization techniques to convert blendshape data into motor control values. Compared to traditional methods, neural network–based approaches can leverage large-scale datasets to automatically learn diverse expression features, offering improved adaptability and scalability~\cite{wu2024retargeting}.

More recently, diffusion models have gained attention as generative models that use a gradual denoising process. Initially proposed by Sohl-Dickstein~\cite{sohl2015deep}, these models have recently gained attention and have been applied in various robotics domains, including reinforcement learning~\cite{ho2020denoising}, imitation learning~\cite{zhu2023diffusion}, and motion planning~\cite{serifi2024robot}. Their core principle is the reverse diffusion process, where data is iteratively denoised to recover target information, effectively capturing temporal dependencies in time-series data~\cite{song2020score}.

Despite their potential, diffusion models have seen limited application in the area of robotic facial retargeting. Previous neural network methods often require long processing times and yield suboptimal results for dynamic control tasks~\cite{chen2021smile,he2016identity}. In this study, we introduce ExFace, a Diffusion Transformer–based system that achieves lower errors and more consistent visual results, while effectively handling both real-time and sequential data across various robotic platforms.

\section{Method}

The goal of this work is to map human facial expressions, represented by Blendshape~\cite{joshi2006learning} data, to the motor control values of a robot’s face, enabling the robot to mimic human-like facial movements. Our network takes as input a sequence of Blendshape values that capture the dynamic motion of key facial features, including the eyes, mouth, and eyebrows. The output consists of motor control values that drive the robot’s actuators, allowing it to reproduce these expressions with precision. By leveraging Diffusion Transformers, our approach effectively captures and generates even complex, rapidly changing facial expressions in real time, ensuring both accuracy and fluidity.

The following sections detail the key components of our method: \emph{Data Collection}, \emph{Network Structure}, and the \emph{Model Bootstrap} strategy, each playing a crucial role in enabling expressive and lifelike robotic facial movements.

\subsection{Data Collection}

\begin{figure}[]
\centering
\includegraphics[width=3in]{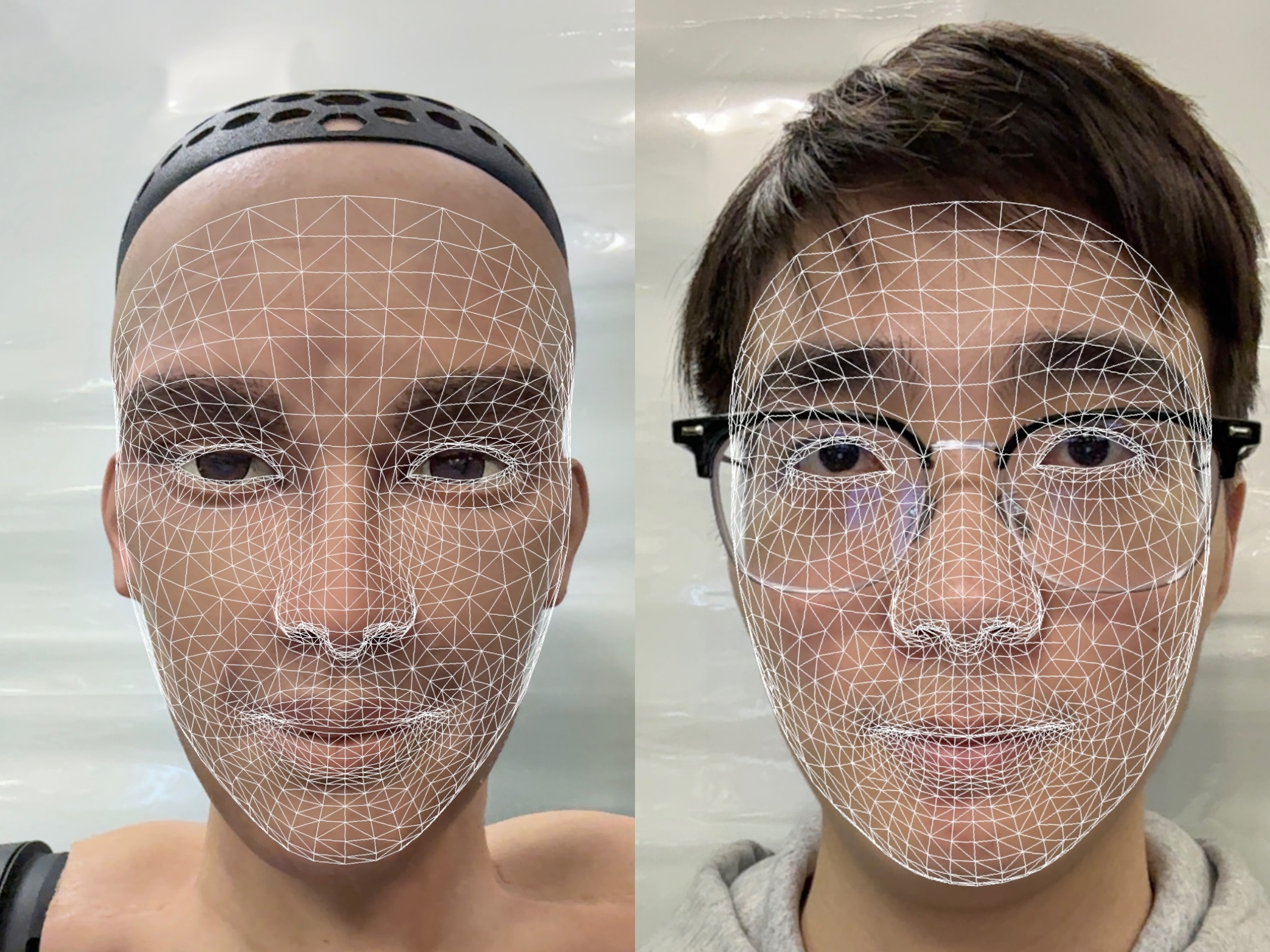}
\caption{Data collection. ARKit~\cite{apple2025arkit} accurately captures blendshape~\cite{joshi2006learning} data from both human and robot faces, providing a reliable foundation for subsequent calibration and mapping.}
\label{fig_2}
\end{figure}

\begin{figure*}[!t]
\centering
\includegraphics[width=7in]{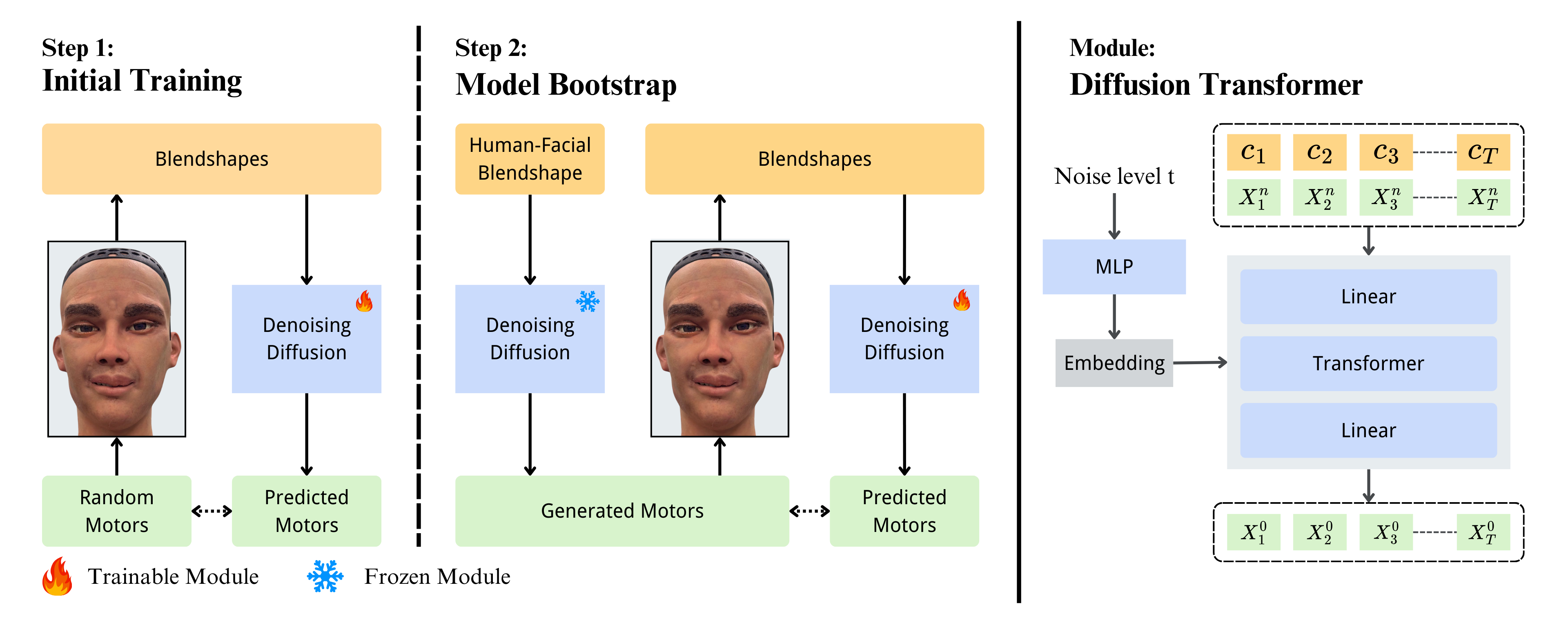}
\caption{Overview of ExFace. ExFace is a system that maps human facial Blendshape data to robot motor control values using a Diffusion Transformer network. On the left side, we begin with initial training using random motor control values, followed by model bootstrap, where dynamic facial expression sequences are used to iteratively refine the model. On the right side, our transformer diffusion network architecture takes Blendshape data (\(c\)) as a condition input. In each iteration, the network progressively denoises the data, and the final output, \(x_0\), is the motor control values.}
\label{fig_3}
\end{figure*}

The experiments presented in this work were conducted on Micheal, the realistic humanoid robot~\cite{hanson2025sophia}, which is equipped with 42 degrees of freedom (DOF) in its head. Given that the primary focus of this research is on facial movements, the degrees of freedom associated with head motion and eye rotation were excluded from the analysis. Instead, we concentrated on a subset of 33 DOF that specifically control facial expressions. This approach allows a more precise examination of the dynamic changes in facial movements, which are critical to the objectives of the study.

Our data collection phase primarily relies on Apple’s ARKit technology shown in Fig.\ref{fig_2}~\cite{apple2025arkit}, which accurately captures facial expressions and provides robust, generalizable blendshape data. To acquire facial expression data, we mounted an iPad in front of the Micheal robot and used ARKit to track and record facial movements. The ARKit output comprises blendshape data that encodes the motion of 55 key facial landmarks, effectively capturing the intricate details of human expressions.

During the data acquisition process, we first perform a calibration step by capturing an image of a neutral facial expression, which will serve as the reference for all subsequent data collection. ARKit then captures raw blendshape data, which is adjusted relative to the neutral expression, generating calibrated blendshape values. Due to ARKit's generalizability, the system can be adapted to different individuals by adjusting to each person's unique calibration.

\subsection{Network Structure}

The core of our ExFace system, shown in Fig.\ref{fig_3}, is a Diffusion Transformer~\cite{peebles2023scalable} architecture that efficiently maps human facial blendshape sequences to bionic robot motor control values. The system is designed to process continuous facial expression sequences of 120 frames, ensuring real-time responsiveness for dynamic facial movements. Inspired by the recent successes of diffusion models in image generation~\cite{chi2023diffusion} and 3D human motion estimation~\cite{li2023ego}, we propose a conditional diffusion model-based framework. Our system maps 55-dimensional blendshape sequences to 33-dimensional motor control values using an encoder-decoder structure that incorporates the principles of denoising diffusion probabilistic models (DDPM)~\cite{ho2020denoising} for progressive noise removal and signal recovery.

To formalize our model within the diffusion framework, we denote \( x_n \) as the motor control sequence at noise level \( n \), comprising \( T \) frames, \( x_n = \{X_n^1, X_n^2, ..., X_n^T\} \). The original, noise-free motor control sequence is represented by \( x_0 \), and our objective is to map \( x_0 \) to the corresponding robot motor control values.

The diffusion model defines a forward process where Gaussian noise is progressively added to \( x_0 \) over \( N \) steps via a Markov chain:
\begin{equation}\label{eq1}
q(x_{1:N} \mid x_0) := \prod_{n=1}^{N} q(x_n \mid x_{n-1}) 
\end{equation}

At each step, the noise is determined by a variance schedule \( \beta_n \), given by:
\begin{equation}\label{eq2}
q(x_n \mid x_{n-1}) := \mathcal{N}\left(x_n; \sqrt{1-\beta_n}\, x_{n-1}, \beta_n I\right)
\end{equation}

After \( N \) steps, \( x_N \) approximates a sample from a standard normal distribution. To generate motor control values conditioned on the blendshapes, we employ a reverse diffusion process, approximated by a Markov chain with learned means and fixed variances:
\begin{equation}\label{eq3}
p_\theta(x_{n-1} \mid x_n, c) := \mathcal{N}\left(x_{n-1}; \mu_\theta(x_n, n, c), \sigma_n^2 I\right) 
\end{equation}
Here, \( \theta \) denotes the network parameters and \( c \) the conditioning input (i.e., the blendshape). The learned mean \( \mu_\theta(x_n, n, c) \) is modeled as:
\begin{equation}\label{eq4}
    \mu_\theta = \frac{\sqrt{\alpha_n}(1 - \bar{\alpha}_{n-1})\, x_n + \sqrt{\bar{\alpha}_{n-1}}(1 - \alpha_n)\, \hat{x}_\theta(x_n, n, c)}{1 - \bar{\alpha}_n}.
\end{equation}

The loss function is calculated using the mean squared error (MSE) between the predicted motor control values and the actual target motor control values, allowing the network to minimize the difference between the generated and real expressions.

\subsection{Model Bootstrap}

Our model bootstrap training strategy consists of multiple iterative stages, progressively improving the model's performance by refining the mapping between motor control values and blendshape data. Initially, we apply random single-frame motor signals to collect static robot blendshape data. This first stage helps to establish sparse sampling of the mapping space. At this stage, the model learns to map these blendshape data to motor control values, and the initial mapping space is scattered and imprecise.

Next, we introduce dynamic facial expression sequences for training. We first capture a segment of facial expression sequences and use this preliminary model to generate motor control values. The robot then performs these expressions and captures the generated robot facial dynamic sequence blendshape data. This new robot-generated blendshape data, along with the corresponding motor control values, is fed back into the network to improve the model’s accuracy and expand its sequence inference capability. Through this iterative process, the model gradually transitions from sparse initial sampling to denser sampling, focusing on the human expression domain.

We use sequences of 120 frames as input/output for the model, which significantly improves the quality of expression generation. To initialize the model, we use 600 pairs of blendshape and motor control values, which are then replicated across 72,000 frames to train the model in the first stage. In the first iteration, we add 8,000 frames of data. Afterward, we add 4,000 frames of data with each iteration for further training. In this dynamic phase, we use the current model to generate motor control values from human facial data and then collect the corresponding robot blendshape data. These motor and blendshape data are used to fine-tune the model, continuing the iterative process until the model achieves high-quality performance.

Through this iterative approach, the system gradually improves its ability to generate smooth, accurate, and stable facial expressions, ultimately enabling the model to effectively handle dynamic driving conditions.

\section{Experiment}

In this section, we comprehensively evaluate the performance of the proposed ExFace system for facial expression retargeting and compare it with baseline models (MLP and Transformer), which are trained on the same data. We assess various quantitative metrics to validate the advantages of ExFace in motor control precision, blendshape generation quality, and real-time responsiveness. We also showcase the effectiveness of our bootstrap training strategy and the cross-platform adaptability of our system.
To validate the performance of ExFace, we have created a 2000-frame complex human natural expression sequence as a validation set. This sequence includes various expressions, speeds, and the application of different muscle groups, aiming to evaluate ExFace's performance in handling more complex dynamic expression changes.

First, we adopt motor distance, measured by mean squared error (MSE), as an evaluation metric to quantify the discrepancy between the predicted motor control values and the ground truth data. Experimental results show that ExFace achieves significantly lower motor loss than the baseline methods, indicating that the combination of diffusion-based denoising and Transformer temporal modeling enables a more accurate capture of the complex mapping between human facial blendshapes and robot motor control values.

Next, to assess the practical effectiveness of facial expression retargeting, we collected blendshape data generated from the robot after applying the predicted motor control values and computed the blendshape distance (MSE between predicted and actual blendshapes). The results shown in Table.\ref{tab:comparison} demonstrate that ExFace produces blendshape outputs that closely match the actual expressions, whereas the traditional models exhibit larger errors, resulting in discontinuous and jittery expression generation.

\begin{table}[H]
\centering
\caption{{Comparison for Different Methods.}}
\label{tab:comparison}
\begin{tabular}{c|c|c}
\toprule
\textbf{Method} & \textbf{Motor Distance} & \textbf{Blendshape Distance}\\
\midrule
\textbf{Random} &0.1461&0.0105\\
\textbf{MLP} &0.0465&0.0039\\
\textbf{Transformer} &0.0383&0.0029\\
\textbf{Ours} &0.0353&0.0025\\
\bottomrule
\end{tabular}
\end{table}

\begin{figure}[H]
\centering
\includegraphics[width=2.7in]{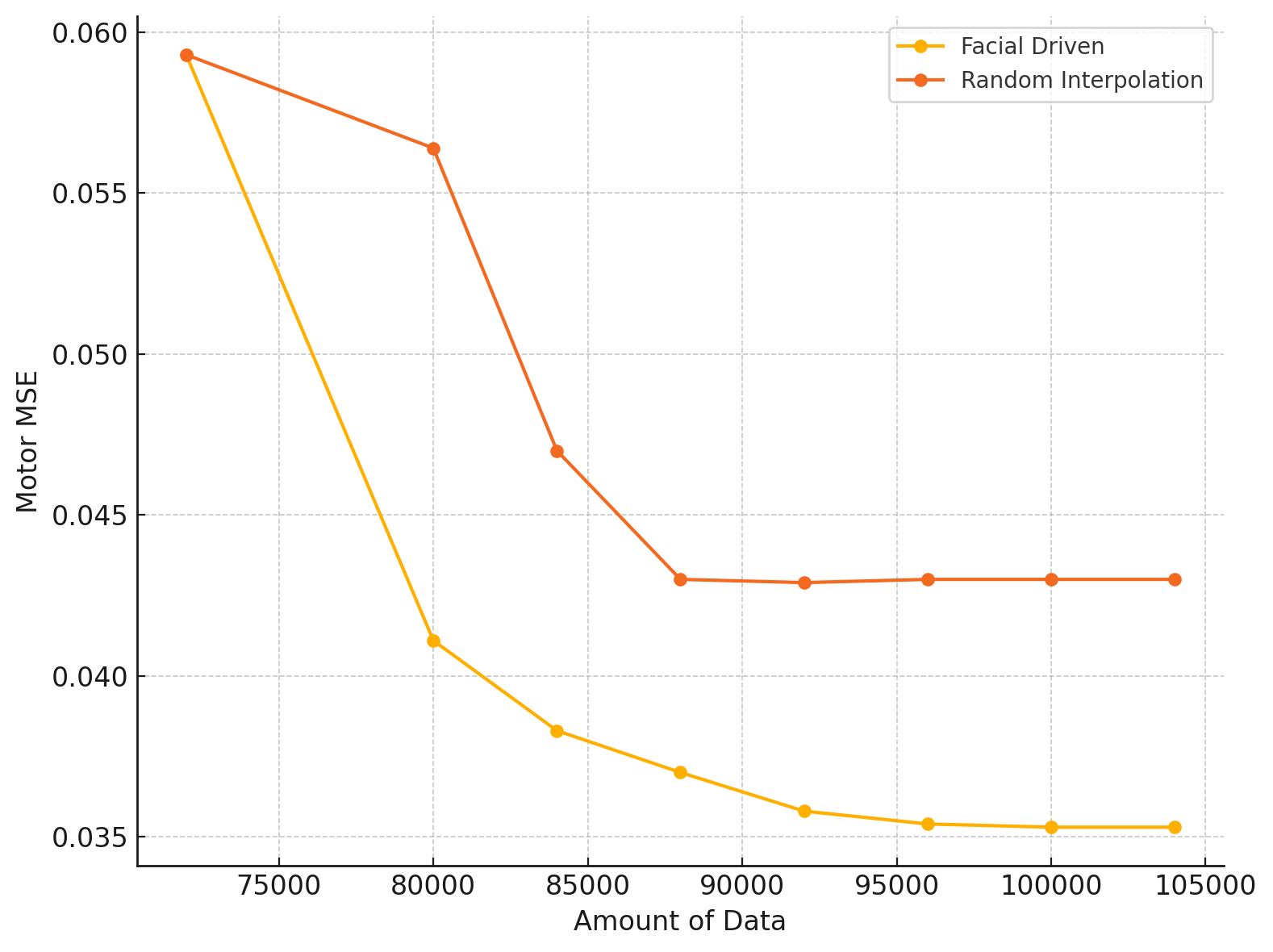}
\caption{ExFace Model Bootstrap Curve. This figure illustrates the performance improvement achieved through our iterative training strategy. It also compares the results of using random data interpolation with dynamic facial expression sequences.}
\label{fig:motor_mse}
\label{fig_4}
\end{figure}

We also evaluated the effect of model bootstrap training by comparing the training results under different amounts of data. Starting from 72,000 frames of static random data (600 frames), we initially added 8,000 frames. Then, we added 4,000 frames of data with each iteration for further training. With each iteration, ExFace generates increasingly natural facial expressions, closer to natural human expressions. At the same time, we trained iterative dynamic sequences based on random data interpolation. Experimental results shown in Fig.\ref{fig_4} show that training with facial-driven dynamic expression sequence data significantly outperforms training with random data interpolation, generating more natural facial expressions. 

To validate the real-time performance of ExFace, we established a TCP/IP-based real-time control system over WiFi in our laboratory. After calibration, continuous control was implemented on the robot workstation: the ARKit app extracted blendshape values in real time from the user’s face and, via the trained network, converted them into motor control commands for the robot Micheal. To ensure real-time responsiveness and accurate data transmission, the ARKit app’s data cache was cleared at the start of each control cycle to prevent delays due to cache accumulation. Empirical measurements indicate that ExFace achieves an overall processing delay of approximately 0.15 seconds, effectively reducing output jitter and ensuring smooth, natural facial expression transitions during real-time interaction.

\begin{figure}[!t]
\centering
\includegraphics[width=3.3in]{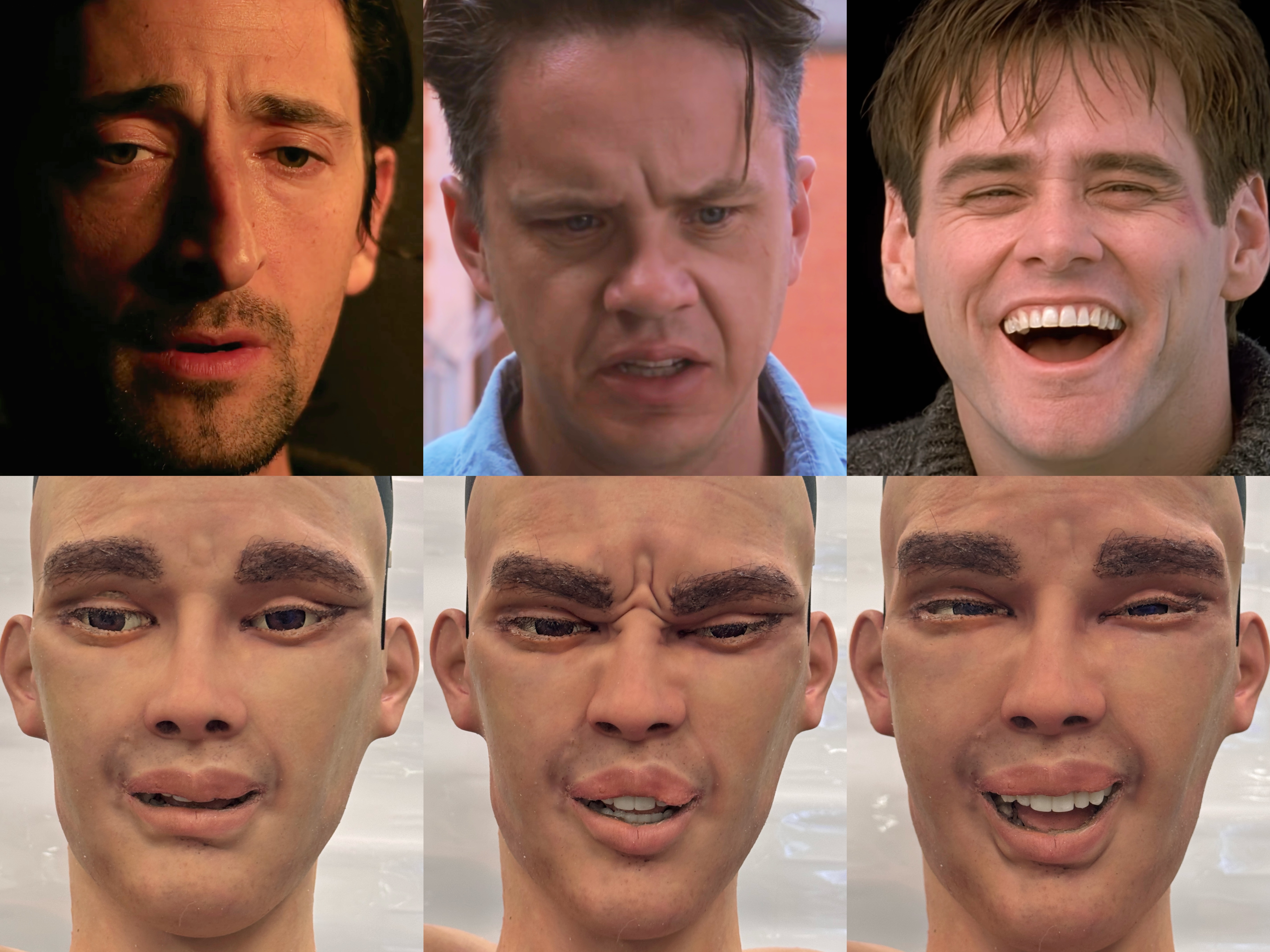}
\caption{Movie-Driven Robot Performance. This figure shows a robot performance driven by a movie segment. The robot accurately replicates the actor's facial expressions.}
\label{fig_6}
\end{figure}

Finally, we tested the model on two distinct robots: Micheal, which utilizes cable-driven actuation with 33 DOF, and Hobbs, which operates with articulated linkages and 32 DOF. The two robots also differ significantly in their actuation points and configurations. We evaluated the model using real-time facial motion, and the results are shown in Fig.\ref{fig_6}. The results indicate that ExFace generates accurate control values on both robots, highlighting the strong applicability of our model across different platforms.

\begin{figure}[!t]
\centering
\includegraphics[width=3.3in]{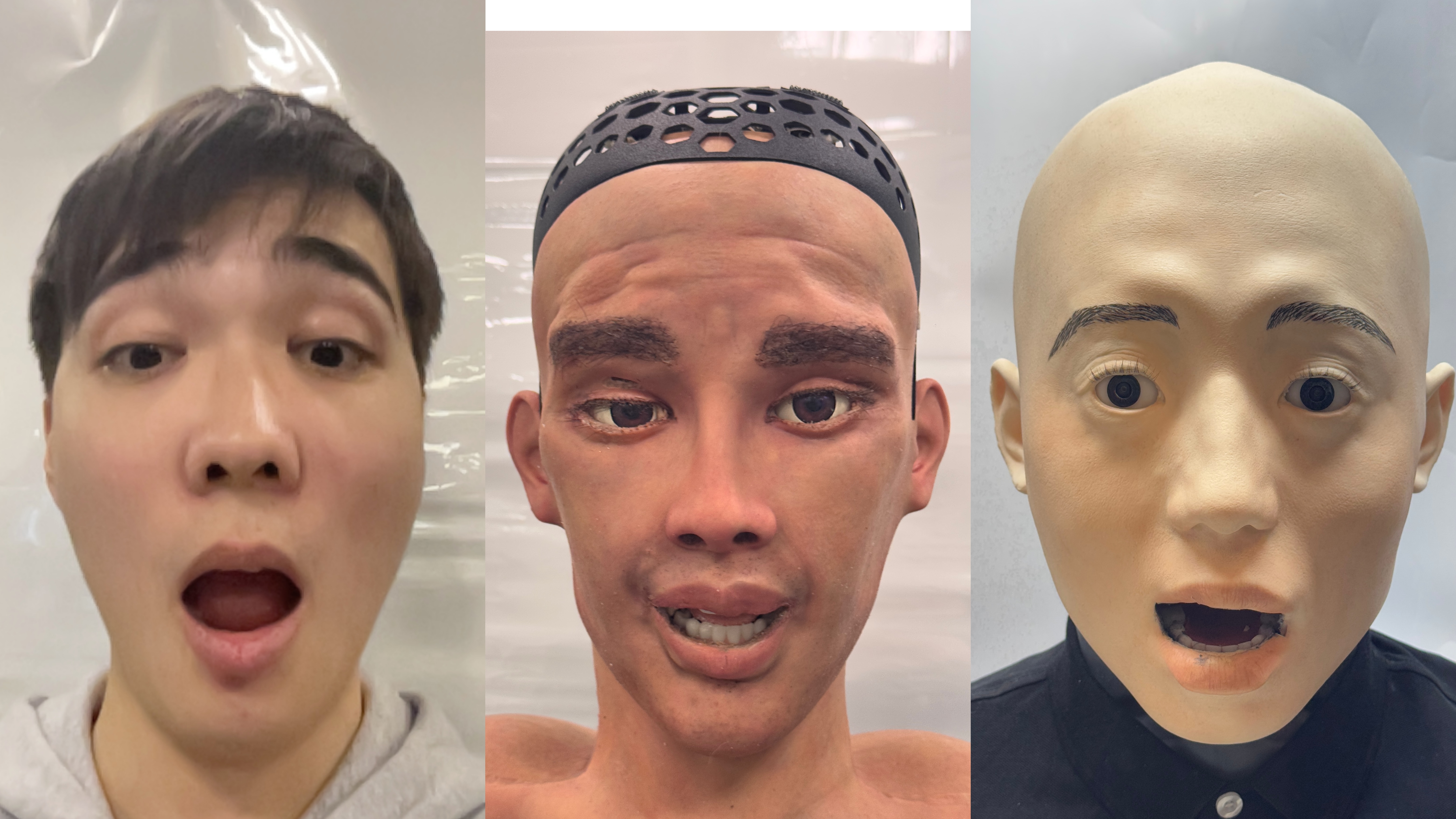}
\caption{Real Person-Driven Performance on Different Robots. This figure demonstrates the generalizability of our approach by showcasing real human-driven facial expressions across various robotic platforms. }
\label{fig_5}
\end{figure}

In summary, ExFace outperforms traditional optimization-based methods in terms of motor loss, blendshape loss, data efficiency, and real-time performance. Our experiments validate its promising application prospects in achieving high-quality, real-time, and cross-platform human-robot interaction.


\section{Application}

\subsection{Real-Time Facial Expression Control}
Real-time facial expression control plays a crucial role in the ExFace system, especially in scenarios involving robotic mimicry and performance. By leveraging the trained neural network, the system precisely converts human facial blendshape data into motor control values, enabling the robot to replicate specific facial expressions in real time. This technology is capable not only of reproducing simple expressions but also of handling more complex applications, such as robotic performances and the reenactment of classic scenes from films or theater, as shown in Fig. \ref{fig_5}. With accurate input into the network, the robot can display every subtle detail of facial motion, resulting in more vivid, expressive, and emotionally engaging performances.

\subsection{Digital Human Interaction}
ExFace also incorporates Media2Face technology~\cite{zhao2024media2face} to enable digital humans to generate corresponding facial expressions based on audio inputs (such as music or speech) or textual cues (such as emotional descriptions). The blendshape data produced in this process can be transmitted in real time to the robot, allowing it to engage in interactive behaviors like singing along with music or delivering expressive presentations. Moreover, by integrating Media2Face with Azure’s speech-to-text capabilities and GPT models, ExFace facilitates real-time human-robot dialogue. In these interactions, the robot not only communicates verbally but also conveys emotions through its facial expressions, leading to a more natural and dynamic conversational experience. This approach effectively transfers digital human interaction techniques to the robotic domain, enabling richer and more nuanced emotional exchanges.


\section{Discussion}

While ExFace has demonstrated significant improvements over traditional methods in facial expression retargeting, several challenges remain. One major issue stems from the reliance on ARKit and Live Link Face applications, which are based on 2D techniques. These technologies often produce suboptimal results, particularly in capturing the nuances of the neck and eye regions, and require well-lit environments, which leads to high noise levels in complex settings. Moreover, as ARKit is primarily trained on human facial data, its performance deteriorates when applied to subjects with poor makeup or bionic robotic faces, making precise expression mapping challenging.

Another challenge is the high latency in digital human generation systems, which take 2–5 seconds to generate results, hindering real-time interaction. Current robotic facial expression control approaches often mimic human expressions without considering the unique features of bionic robots, sometimes resulting in unnatural results. We are exploring ways to integrate bionic robot design features, ensuring generated expressions are more aligned with the robot's facial characteristics.

We are also considering incorporating prior knowledge to reduce dataset requirements. Given the similarities between facial action units (AUs), ARKit features, and motor control values, we aim to improve expression control with less data, reducing the risk of damage to the robot's facial surface.

Finally, our work contributes to the development of digital twins for bionic robots. Most existing bionic robots are designed manually and rely on numerous heuristics. By training our blendshape-to-motor mapping network and generating digital human models of bionic robots from images, we can simulate and optimize facial expression dynamics in a virtual environment. This not only prevents potential damage to the robot's physical structure but also enhances its expressive capabilities by addressing current design limitations.

We also combine the final training dataset with human-driven performances, movie clip-driven data, and digitally generated human expressions for transfer learning. This mixed dataset, released as the ExFace dataset, includes robot facial images, motor control values, and corresponding blendshape data.


\section{Conclusions}

The proposed ExFace system leverages a Diffusion Transformer network to achieve efficient mapping from human facial blendshapes to bionic robot motor control values. By incorporating an innovative model bootstrap training strategy, the system continuously refines expression generation, ensuring that the produced facial movements are both natural and precise. Experimental results demonstrate that ExFace excels in real-time performance and accuracy in facial expression retargeting, successfully meeting the demands for high-quality expression generation. Moreover, its application in robot performances and digital human interaction showcases its broad potential, offering a novel technological pathway for more vivid and authentic human–machine interactions.


\bibliographystyle{unsrt}
\bibliography{references}

\begin{thebibliography}{10}

\bibitem{plutchik2014emotions}
Robert Plutchik.
\newblock Emotions: A general psychoevoiutionary theory.
\newblock In {\em Approaches to emotion}, pages 197--219. Psychology Press, 2014.

\bibitem{fong2003survey}
Terrence Fong, Illah Nourbakhsh, and Kerstin Dautenhahn.
\newblock A survey of socially interactive robots.
\newblock {\em Robotics and autonomous systems}, 42(3-4):143--166, 2003.

\bibitem{reissland1988neonatal}
Nadja Reissland.
\newblock Neonatal imitation in the first hour of life: Observations in rural nepal.
\newblock {\em Developmental Psychology}, 24(4):464, 1988.

\bibitem{hakli2014social}
Raul Hakli.
\newblock Social robots and social interaction.
\newblock In {\em Sociable robots and the future of social relations}, pages 105--114. IOS Press, 2014.

\bibitem{lin2016expressional}
Chyi-Yeu Lin, Chun-Chia Huang, and Li-Chieh Cheng.
\newblock An expressional simplified mechanism in anthropomorphic face robot design.
\newblock {\em Robotica}, 34(3):652--670, 2016.

\bibitem{blow2006art}
Mike Blow, Kerstin Dautenhahn, Andrew Appleby, Chrystopher~L Nehaniv, and David Lee.
\newblock The art of designing robot faces: Dimensions for human-robot interaction.
\newblock In {\em Proceedings of the 1st ACM SIGCHI/SIGART conference on Human-robot interaction}, pages 331--332, 2006.

\bibitem{chen2021smile}
Boyuan Chen, Yuhang Hu, Lianfeng Li, Sara Cummings, and Hod Lipson.
\newblock Smile like you mean it: Driving animatronic robotic face with learned models.
\newblock In {\em 2021 IEEE International Conference on Robotics and Automation (ICRA)}, pages 2739--2746. IEEE, 2021.

\bibitem{wu2024retargeting}
Bowen Wu, Chaoran Liu, Carlos~T Ishi, Takashi Minato, and Hiroshi Ishiguro.
\newblock Retargeting human facial expression to human-like robotic face through neural network surrogate-based optimization.
\newblock In {\em 2024 IEEE/RSJ International Conference on Intelligent Robots and Systems (IROS)}, pages 4724--4730. IEEE, 2024.

\bibitem{joshi2006learning}
Pushkar Joshi, Wen~C Tien, Mathieu Desbrun, and Fr{\'e}d{\'e}ric Pighin.
\newblock Learning controls for blend shape based realistic facial animation.
\newblock In {\em ACM Siggraph 2006 Courses}, pages 17--es. 2006.

\bibitem{peebles2023scalable}
William Peebles and Saining Xie.
\newblock Scalable diffusion models with transformers.
\newblock In {\em Proceedings of the IEEE/CVF international conference on computer vision}, pages 4195--4205, 2023.

\bibitem{ho2020denoising}
Jonathan Ho, Ajay Jain, and Pieter Abbeel.
\newblock Denoising diffusion probabilistic models.
\newblock {\em Advances in neural information processing systems}, 33:6840--6851, 2020.

\bibitem{vaswani2017attention}
Ashish Vaswani, Noam Shazeer, Niki Parmar, Jakob Uszkoreit, Llion Jones, Aidan~N Gomez, {\L}ukasz Kaiser, and Illia Polosukhin.
\newblock Attention is all you need.
\newblock {\em Advances in neural information processing systems}, 30, 2017.

\bibitem{scherer1992does}
K~Scherer.
\newblock What does a facial expression express.
\newblock 1992.

\bibitem{oh2006design}
Jun-Ho Oh, David Hanson, Won-Sup Kim, Young Han, Jung-Yup Kim, and Ill-Woo Park.
\newblock Design of android type humanoid robot albert hubo.
\newblock In {\em 2006 IEEE/RSJ International Conference on Intelligent Robots and Systems}, pages 1428--1433. IEEE, 2006.

\bibitem{asheber2016humanoid}
Wagshum~Techane Asheber, Chyi-Yeu Lin, and Shih~Hsiang Yen.
\newblock Humanoid head face mechanism with expandable facial expressions.
\newblock {\em International Journal of Advanced Robotic Systems}, 13(1):29, 2016.

\bibitem{sohl2015deep}
Jascha Sohl-Dickstein, Eric Weiss, Niru Maheswaranathan, and Surya Ganguli.
\newblock Deep unsupervised learning using nonequilibrium thermodynamics.
\newblock In {\em International conference on machine learning}, pages 2256--2265. pmlr, 2015.

\bibitem{zhu2023diffusion}
Zhengbang Zhu, Hanye Zhao, Haoran He, Yichao Zhong, Shenyu Zhang, Haoquan Guo, Tingting Chen, and Weinan Zhang.
\newblock Diffusion models for reinforcement learning: A survey.
\newblock {\em arXiv preprint arXiv:2311.01223}, 2023.

\bibitem{serifi2024robot}
Agon Serifi, Ruben Grandia, Espen Knoop, Markus Gross, and Moritz B{\"a}cher.
\newblock Robot motion diffusion model: Motion generation for robotic characters.
\newblock In {\em SIGGRAPH Asia 2024 Conference Papers}, pages 1--9, 2024.

\bibitem{song2020score}
Yang Song, Jascha Sohl-Dickstein, Diederik~P Kingma, Abhishek Kumar, Stefano Ermon, and Ben Poole.
\newblock Score-based generative modeling through stochastic differential equations.
\newblock {\em arXiv preprint arXiv:2011.13456}, 2020.

\bibitem{he2016identity}
Kaiming He, Xiangyu Zhang, Shaoqing Ren, and Jian Sun.
\newblock Identity mappings in deep residual networks.
\newblock In {\em Computer Vision--ECCV 2016: 14th European Conference, Amsterdam, The Netherlands, October 11--14, 2016, Proceedings, Part IV 14}, pages 630--645. Springer, 2016.

\bibitem{apple2025arkit}
Apple.
\newblock Apple arkit, 2024.
\newblock Accessed: 2025-02-26.

\bibitem{hanson2025sophia}
Hanson Robotics.
\newblock Sophia, 2025.
\newblock Accessed: 2025-02-27.

\bibitem{chi2023diffusion}
Cheng Chi, Zhenjia Xu, Siyuan Feng, Eric Cousineau, Yilun Du, Benjamin Burchfiel, Russ Tedrake, and Shuran Song.
\newblock Diffusion policy: Visuomotor policy learning via action diffusion.
\newblock {\em The International Journal of Robotics Research}, page 02783649241273668, 2023.

\bibitem{li2023ego}
Jiaman Li, Karen Liu, and Jiajun Wu.
\newblock Ego-body pose estimation via ego-head pose estimation.
\newblock In {\em Proceedings of the IEEE/CVF Conference on Computer Vision and Pattern Recognition}, pages 17142--17151, 2023.

\bibitem{zhao2024media2face}
Qingcheng Zhao, Pengyu Long, Qixuan Zhang, Dafei Qin, Han Liang, Longwen Zhang, Yingliang Zhang, Jingyi Yu, and Lan Xu.
\newblock Media2face: Co-speech facial animation generation with multi-modality guidance.
\newblock In {\em ACM SIGGRAPH 2024 conference papers}, pages 1--13, 2024.

\end{thebibliography}

\end{document}